\documentclass[letterpaper]{article} 
\usepackage{aaai23}  
\usepackage{times}  
\usepackage{helvet}  
\usepackage{courier}  
\usepackage[hyphens]{url}  
\usepackage{graphicx} 
\usepackage{natbib}  
\usepackage{caption} 
\newcommand{\modelname}{PRI-GSL}
\usepackage{amsthm,amsmath,amssymb}
\usepackage{newfloat}
\usepackage{listings}

\usepackage{wrapfig}
\usepackage{subfigure}
\usepackage{enumitem}
\usepackage{pifont}
\usepackage{float}
\usepackage{multirow}
\usepackage[ruled]{algorithm2e}
\usepackage{color}
\usepackage[normalem]{ulem}
\useunder{\uline}{\ul}{}
\usepackage{makecell}

\newtheorem{myDef}{Definition}

\usepackage{newfloat}
\usepackage{listings}
\DeclareCaptionStyle{ruled}{labelfont=normalfont,labelsep=colon,strut=off} 
\floatstyle{ruled}
\newfloat{listing}{tb}{lst}{}
\floatname{listing}{Listing}
\title{Self-organization Preserved Graph Structure Learning\\with Principle of Relevant Information}
\author{
    Qingyun Sun\textsuperscript{\rm 1}\textsuperscript{\rm 2},
    Jianxin Li\textsuperscript{\rm 1}\textsuperscript{\rm 2}, Beining Yang\textsuperscript{\rm 1}\textsuperscript{\rm 2}, Xingcheng Fu\textsuperscript{\rm 1}\textsuperscript{\rm 2}, Hao Peng\textsuperscript{\rm 1}, Philip S. Yu\textsuperscript{\rm 3}\\
}
\affiliations {
    \textsuperscript{\rm 1} Beijing Advanced Innovation Center for Big Data and Brain Computing,
    Beihang University,
    Beijing 100191, China\\
    \textsuperscript{\rm 2}
    School of Computer Science and Engineering, 
    Beihang University, 
    Beijing 100191, China\\
    \textsuperscript{\rm 3} Department of Computer Science, University of Illinois at Chicago, Chicago, USA\\
    \{sunqy, lijx, yangbeining, fuxc, penghao\}@buaa.edu.cn, 
    psyu@uic.edu
}
\usepackage{bibentry}

\begin{document}

\maketitle

\begin{abstract}
Most Graph Neural Networks follow the message-passing paradigm, assuming the observed structure depicts the ground-truth node relationships. 
However, this fundamental assumption cannot always be satisfied, as real-world graphs are always incomplete, noisy, or redundant. 
How to reveal the inherent graph structure in a unified way remains under-explored. 
We proposed \textbf{\modelname}, a \textbf{G}raph \textbf{S}tructure \textbf{L}earning framework guided by the \textbf{P}rinciple of \textbf{R}elevant \textbf{I}nformation, providing a simple and unified framework for identifying the self-organization and revealing the hidden structure. 
\modelname~learns a structure that contains the most relevant yet least redundant information quantified by von Neumann entropy and Quantum Jensen-Shannon divergence. 
\modelname~incorporates the evolution of quantum continuous walk with graph wavelets to encode node structural roles, showing in which way the nodes interplay and self-organize with the graph structure. 
Extensive experiments demonstrate the superior effectiveness and robustness of \modelname. 
\end{abstract}

\section{Introduction}
Graph Neural Networks (GNNs)~\cite{wu2020comprehensive} have gained popularity in recent years due to their remarkable success in representing graph data in diverse tasks and applications. 
Most of the existing GNNs follow the message-passing paradigm~\cite{gilmer2017neural}, i.e., exchanging information between neighbors along the graph structure. 
They take the raw graph structure as the path of information flow, assuming the observed structure perfectly depicts the ground-truth relations between nodes. 
However, these raw graphs are naturally collected from network-structure data (e.g., social networks), which are often noisy, incomplete, and independent of the downstream tasks. 
There is a gap between the raw structure and the optimal structure for specific tasks. 
The poor quality of graph structure leads to the poor quality of representations produced by GNNs, making GNNs prone to noise and adversarial attacks~\cite{zugner2018adversarial,sun2018adversarial,sun2021sugar}. 

Graph structure learning~\cite{zhu2021deep} aims to learn a new structure of high quality simultaneously with the graph representations, which has received growing attention for its utility for improving representation quality and robustness. 
Most existing methods optimize the structure  with heuristic assumptions (e.g., community~\cite{wang2021graph}) or certain structure constraints  (e.g., sparsity, low-rank, and smoothness~\cite{jin2020graph,sun2022position}). 
However, these assumptions and constraints cannot always be applicable to all graphs and tasks. 
\textit{How to reveal the inherent graph structure in a unified way remains an under-explored question}. 


\begin{figure}
    \centering
    \includegraphics[width=\linewidth]{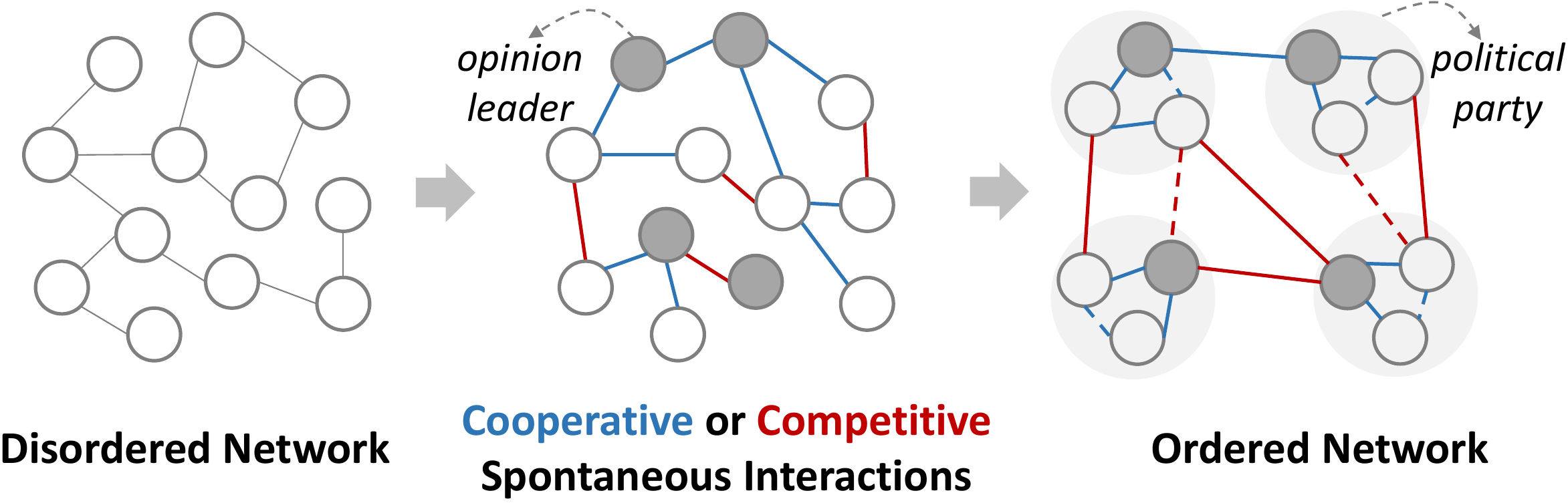}
    \caption{Self-organization in political network. }
    \label{fig:self-organization}
\end{figure}

Most of the graph data in the real-world shows \textit{``self-organization''} property from molecules~\cite{eigen1977principle} to social networks~\cite{bonabeau1997self}, where the nodes organize their interactions spontaneously through the structure to create a global order amongst themselves. 
As the example in Fig.~\ref{fig:self-organization}, the emergence of opinion leaders and the cooperation/competitive behaviors between people form political parties, making the political network more ordered. 
Since the structure navigates the information flow between nodes and decides the graph's fundamental mechanism, it can be optimized to identify the organization and reduce the disorder of the noisy graph.


In this paper, we introduce the self-organized Principle of Relevant Information (PRI) to quantify the structure from an information-theoretic point of view. 
We propose a novel graph structure learning framework named \modelname, 
which inherits the merits of PRI to identify the self-organization and reveal the inherent structure of graph. 
Rather than imposing statistical constraints on the graph data, \modelname~takes the structure learning as a trade-off between structure redundancy reduction and information preservation, and then use the von Neumann entropy and the Quantum Jensen-Shannon divergence to quantify them. 
To better capture the contribution of nodes in the self-organization evolution process, we use the quantum continuous walk evolution with multi-scale graph wavelets to characterize node structural roles and incorporate them into structure learning. 
In this way, \modelname~enumerates the potential edges and preserves the most relevant yet least redundant ones, 
showing in which way the nodes interplay and self-organize with the graph structure. 

\begin{itemize}[leftmargin=*]
    \item We propose \modelname, an information-theoretic graph structure learning framework with the Principle of Relevant Information, providing a simple yet unified way to quantify the learned structure and unravel the graph self-organization. 
    \item 
    We use the quantum continuous walk with graph wavelets to encode node structural roles in a continuous and time-varying way, which is incorporated in structure learning to fully characterize the nodes in self-organization. 
    \item 
    Extensive experiment results demonstrate the superior effectiveness and robustness of \modelname. 
\end{itemize}
\section{Related Work}
\textbf{Graph structure learning} has gained more attention in recent years~\cite{zhu2021deep} to improve the quality of graph representations by learning a better graph structure. 
Most existing works~\cite{jin2020graph,wang2021graph} optimize the structure with assumptions or certain constraints in a heuristic way. 
Substantial efforts have been made to give a theoretical quantification for the learned structure. 
SDRF~\cite{topping2021understanding}
refines the structure based on the Ricci curvature of edges in a greedy pre-process strategy. 
SIB~\cite{yu2020graph} utilizes the information bottleneck principle to find the most predictive subgraph. 
VIB-GSL~\cite{sun2022graph} proposes a variational information bottleneck principle to learn a new structure for graph classification, which is not applicable to the node-level tasks. 
Graph-PRI~\cite{yu2022principle} advances the Principle of Relevant Information for graph sparsification in an unsupervised way without considering node features and the specific downstream task. 

\noindent\textbf{Information theory} provides a powerful methodology to describe general properties of arbitrarily complex systems. 
In information theory, there are two representative self-organizing principles: Information Bottleneck (IB)~\cite{tishby2000information} and Principle of Relevant Information (PRI)~\cite{principe2010information}. 
Both IB and PRI describe different forms of redundancy reduction and information preservation. 
The famed IB is formulated on the mutual information between independent and identically distributed (i.i.d.) data, which is difficult to model the complex node interactions imposed by the graph structure. 
PRI shares the spirit of the IB method but its formulation addresses the entropy and relative entropy of a single dataset~\cite{principe2010information}, which can be applied to graph data with well-defined information-theoretic tools. 

\section{Preliminary}
\subsubsection{Notions}
Given a graph $G=\{V, E\}$ where $V$ is the set of $N$ nodes and $E$ is the edge set. 
$\mathbf{A}\in \mathbb{R}^{N\times N}$ is the adjacency matrix and $\mathbf{D}$ is the degree matrix. 
The Laplacian matrix of the graph $G$ can be defined as $\mathbf{L}=\mathbf{D}-\mathbf{A} =\mathbf{U}\mathbf{\Lambda }\mathbf{U}^{\mathrm{T}}$, where $\mathbf{U}$ is the eigenvector matrix, $\mathbf{\Lambda }={\rm Diag}(\lambda_1,\cdots,\lambda_N)$ and $\lambda_1 < \lambda_2 \le \cdots \le \lambda_N$ are the eigenvalues of $\mathbf{L}$. 

\subsubsection{Graph Structure Learning}
Given a graph $G$, graph structure learning~\cite{zhu2021deep} aims to learn a new structure $\Tilde{G}$ simultaneously with the graph representations with the objective function: 
\begin{equation}
    \mathcal{L}= \mathcal{L}_{task}(\Tilde{G},Y)+ \alpha\mathcal{L}_{reg}(\Tilde{G}, G),
\end{equation}
where $\mathcal{L}_{task}$ is the task-specific objective with respect to the learned graph $\Tilde{G}$ and the ground truth $Y$, $\mathcal{L}_{reg}$ imposes constraints on the learned graph and $\alpha$ is a hyper-parameter. 


\subsubsection{Principle of Relevant Information}
PRI~\cite{principe2010information} 
is a self-organized information-theoretic principle that aims to perform mode decomposition of a random variable to obtain a reduced statistical representation. 
PRI formulates the redundancy reduction and information preservation as a trade-off between the entropy of reduced representation and its relative entropy given the original data. 
\begin{myDef}[Principle of Relevant Information]
Given a random variable $X$, the Principle of Relevant Information aims to obtain a reduced representation $T$ with: 
\begin{equation}
    \mathcal{L}_{\rm PRI} = \arg\min_{T}H(T)+\beta D(\mathbb{P}(T)||\mathbb{P}(X)),
\end{equation}
where $H(T)$ is the entropy of $T$ and $D(\mathbb{P}(T)||\mathbb{P}(X))$ is the divergence of distributions $\mathbb{P}(T)$ and $\mathbb{P}(X)$. 
\end{myDef}
The first term $H(T)$ measures the redundancy of representation $T$ and the second term $D(\mathbb{P}(T)||\mathbb{P}(X))$ measures the allowable distortion of the original data. 
The hyper-parameter $\beta$ controls the level of distortion in $T$. 
PRI was commonly used in scalar random variables~\cite{wei2021multiscale,hoyos2021relevant} and defined by Rényi's formulation of entropy and divergence~\cite{renyi1961measures}.

\section{Graph Structure Learning with Principle of Relevant Information}
\begin{figure*}
    \centering
    \includegraphics[width=\linewidth]{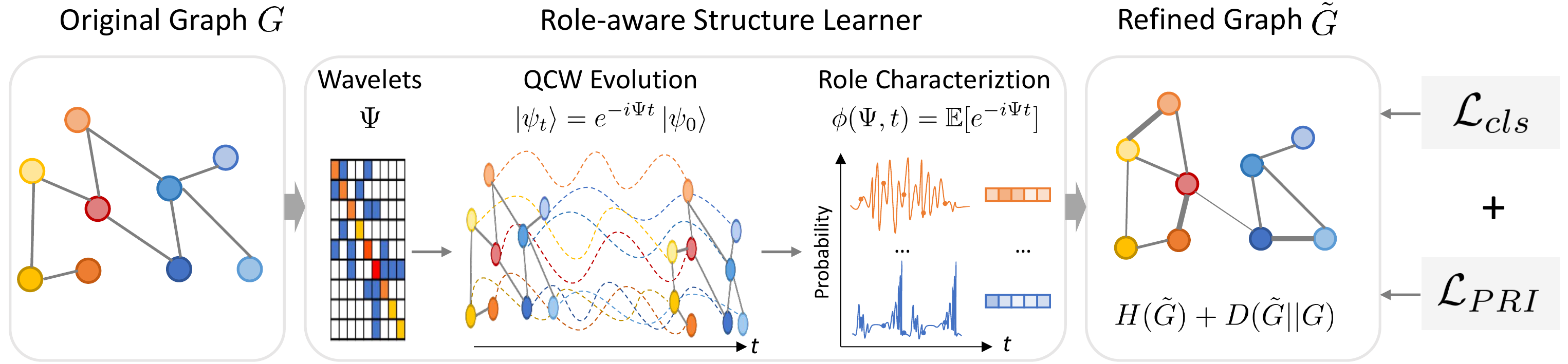}
    \caption{Overall Architecture of \modelname. }
    \label{fig:framework}
\end{figure*}
In this work, we propose a graph structure learning framework named \modelname, which merits the Principle of Relevant Information as a guideline for controlling the structure quality. 
The overall architecture of \modelname~is shown in Figure~\ref{fig:framework}. 
In this section, we first formulate the PRI loss for structure learning, then introduce the Role-aware graph learner and learning process of \modelname. 
\subsection{PRI for Graph Structure Learning}
In the \modelname~framework, PRI performs as a self-supervised regularizer for the quality of the learned graph structure. 
Motivates by the objective of PRI, the Graph Structure Learning Principle of Relevant Information is: 
\begin{myDef}[PRI for Graph Structure Learning]
Given a graph $G$, the Principle of Relevant Information for graph structure learning aims to learn a refined graph $\Tilde{G}$ with:
\begin{equation}
    \mathcal{L}_{\rm PRI}=H(\Tilde{G})+\beta D(\Tilde{G}||G),
\end{equation}
\end{myDef}
The first term $H(\Tilde{G})$ is the redundancy term, which measures the disorder of the learned graph $\Tilde{G}$. 
The larger the $H(\Tilde{G})$, the more disordered $\Tilde{G}$ is. 
The second term $D(\Tilde{G}||G)$ is the distortion term, which measures the discrepancy between two graphs. 
The smaller the $D(\Tilde{G}||G)$, the more similar are the distributions of $\Tilde{G}$ and $G$. 
$\beta$ denotes the trade-off between the redundancy reduction of $\Tilde{G}$ and its discriminative description power of $G$. 
As $\beta$ becomes larger, the emphasis is laid more on the distortion term, and more information from $G$ is preserved in $\Tilde{G}$. 
\subsection{Formulate PRI by von Neumann Entropy and Quantum Jensen-Shannon Divergence}
The choice of entropy and divergence in PRI is application-specific. 
In this paper, we formulate the PRI loss $\mathcal{L}_{\rm PRI}$ by the von Neumann entropy (VNE)~\cite{nielsen2002quantum} and quantum Jensen-Shannon (QJS) divergence~\cite{lamberti2008metric} for graph data with complex interactions as in~\cite{yu2022principle}.


For the first redundancy term $H(\Tilde{G})$, we propose to measure the structure redundancy by von Neumann entropy (VNE)~\cite{nielsen2002quantum}, which has been used in a variety of graph learning studies~\cite{dasoulas2020ego,yu2022principle}. 
von Neumann entropy quantifies the spectral complexity (or disorder) of graph structure by taking the graph as a quantum system through a mapping between discrete Laplacians and quantum states. 
A density matrix $\rho$ is a Hermitian and positive semi-definite matrix that is used to encode the probability distributions and describe the state of a quantum mechanical system. 
For the graph $\Tilde{G}$, the von Neumann entropy $H_{\rm vN}(\Tilde{G})$ is defined as: 
\begin{equation}
\label{eq:VNE}
    H_{\rm vN}(\Tilde{G})=H_{\rm vN}(\Tilde{\rho})
    =-\textbf{tr}\left(\Tilde{\rho}\log \Tilde{\rho}\right)
    =-\sum_{i=1}^{N}\left (\lambda_{i}\log \lambda_{i}\right),  
\end{equation}
where $\Tilde{\rho}$ is the graph density matrix of $\Tilde{G}$, $\textbf{tr}(\cdot)$ denotes trace and $\{\lambda_i\}$ are the eigenvalues of $\Tilde{\rho}$. 
Typically, both the Laplacian matrix and the normalized Laplacian matrix can be used for the mapping from graphs to states~\cite{minello2019neumann}. 
We define the density matrix $\Tilde{\rho}=\frac{\Tilde{\mathbf{L}}}{\textbf{tr}(\Tilde{\mathbf{L}})}=\frac{\Tilde{\mathbf{L}}}{2|E|}$ based on the Laplacian matrix $\Tilde{\mathbf{L}}$ of $\Tilde{G}$, which models the continuous information diffusion process~\cite{de2016spectral}. 

For the second distortion term $D(\Tilde{G}||G)$, we use the quantum Jensen-Shannon (QJS) divergence~\cite{lamberti2008metric} between the graph density matrices ($\Tilde{\rho}$ and $\rho$)  of $G$ and $\Tilde{G}$. 
The quantum Jensen-Shannon divergence has been widely used as a generalization of the classical Jensen-Shannon divergence to quantum states of graph data~\cite{de2015structural,bai2015quantum}, which is 
symmetric, negative definite and bounded ($0 \le D_{\rm QJS} \le 1$). 
\begin{equation}
\label{eq:QJS}
    D_{\rm QJS}(\Tilde{G}||G)=H_{\rm vN}\left(\frac{\Tilde{\rho}+\rho}{2}\right)-\frac{1}{2}H_{\rm vN}\left(\Tilde{\rho}\right)-\frac{1}{2}H_{\rm vN}\left(\mathbf{\rho}\right).
\end{equation}

Combining the redundancy term in Eq.~\eqref{eq:VNE} and the distortion term in Eq.~\eqref{eq:QJS}, we can obtain the following objective: 
\begin{equation}
\label{eq:PRI_loss}
\begin{aligned}
    \mathcal{L}_{\rm PRI}
    &=H_{\rm vN}(\Tilde{G})+\beta D_{\rm QJS}(\Tilde{G}||G)\\
    &=\beta H_{\rm vN}\left(\frac{\Tilde{\rho}+\mathbf{\rho}}{2}\right)+\frac{2-\beta}{2}H_{\rm vN}\left(\Tilde{\rho}\right)-\frac{\beta}{2}H_{\rm vN}\left(\mathbf{\rho}\right)\\
    &\equiv \beta H_{\rm vN}\left(\frac{\Tilde{\rho}+\mathbf{\rho}}{2}\right)+\frac{2-\beta}{2}H_{\rm vN}\left(\Tilde{\rho}\right)
    . 
\end{aligned}
\end{equation}
We neglect $H_{\rm vN}\left(\rho\right)$ in the last line because it's a constant value during optimization. 
The above formalism provides a unified way of quantification and comparison for the learned structure. 
Then we use a graph structure learner to obtain $\Tilde{G}$. 

\subsection{Role-aware Graph Structure Learner}
\label{sec:structure_learner}
In this section, we introduce the Role-aware Graph Structure Learner in \modelname, which aims to learn a better graph $\Tilde{G}$ that preserves the graph self-organization. 
Considering the evolution process in self-organization, we characterize the nodes' roles in a continuous and time-varying way and then incorporate both the merits of features as well as structural roles to refine the graph. 


\subsubsection{Structural Role Encoding}




The structural role of the node represents its contribution to the overall information flow of the graph, which can provide key insights into the identification of graph organization. 
We propose to model graph state by quantum continuous walk and use the time-evolution operator with graph wavelets to generate role encodings. 

\textbf{(1) Model graph state by QCW. } 
Recall that we apply the von Neumann Entropy and the QJS divergence for PRI formulation, which takes the whole graph as a quantum system. 
To investigate the nodes' roles in this quantum system, we use the quantum continuous walk (QCW)~\cite{childs2010relationship,bai2015quantum} to build maps of how information flows through the graph in the perspective of graph state evolution. 
QCW is the quantum mechanical counterpart of the continuous-time random walk in a graph, which describes the propagation of a quantum particle evolving continuously in time on the nodes. 
The QCW on a graph $G$ is defined as a dynamical process over the nodes that takes place on a $N$-dimensional Hilbert space $\mathbb{H}={\rm span}(\{\left | v  \right \rangle,v\in V\})$. 
The evolution of the walker is governed by the Schrödinger equation
\begin{equation}
\label{eq:Schro}
    \frac{d}{dt}\left | \psi_t  \right \rangle =-i\mathcal{H}\left | \psi_t  \right \rangle . 
\end{equation}
$\left | \psi_t  \right \rangle $ represents the state of the walk at time $t$, which is a time-dependent amplitude vector on nodes. 
$\mathcal{H}$ is the Hamiltonian operator, which accounts for the total energy of the graph and governs the time evolution of the quantum continuous walk. 
Given an initial state $\left | \psi_0  \right \rangle\in\mathbb{H}$, the state of walker $\left | \psi_t  \right \rangle$ evolves in time according to
\begin{equation}
\label{eq:evolution}
    \left | \psi_t  \right \rangle =U(t)\left | \psi_0  \right \rangle ,~~~ U(t):=e^{-i\mathcal{H}t},
\end{equation}
where $U(t)$ is the unitary time-evolution operator. 

\textbf{(2) Diffusion by graph wavelets. } 
To give a full characterization of the structural properties, we use the spectral graph wavelets $\Psi$~\cite{hammond2011wavelets} as the Hamiltonian $\mathcal{H}$ in QCW. 
Then the structural information residing in the graph is encoded in $U(t)$. 
$U(t)$ is a polynomial in $\Psi$ for all $t$, thus any matrix that commutes with $\Psi$ also commutes with $U(t)$~\cite{coutinho2021graph}. 
We adopt the heat kernel $g_s(\mathbf{\Lambda })=e^{-\mathbf{\Lambda } s}$ to obtain spectral graph wavelets, 
where the scaling parameter $s$ controls the spread radii of the diffusion process and larger $s$ allows farther diffusion. 
The spectral wavelet basis is 
\begin{equation}
\label{eq:wavelet_s}
    \Psi_s=\mathbf{U}\mathbf{\Lambda }_s\mathbf{U}^{\mathrm{T}}=\left(\Psi_s\left(1\right)|\Psi_s\left(2\right)|\cdots|\Psi_s\left(N\right)\right),
\end{equation}
where $\mathbf{\Lambda }_s=g_s(\mathbf{\Lambda })$. 
In this way, the spectral graph wavelet $\Psi_s(a)$ centered at node $v_a$ associated with filter $g_s$ will be given by an $N$-dimensional vector: 
\begin{equation}
\label{eq:wavelet_s_a}
    \Psi_s(a)=\mathbf{U}\mathbf{\Lambda }_s\mathbf{U}^{\mathrm{T}}\delta_a,
\end{equation}
where $\delta_a$ is the one-hot vector of node $v_a$. 
$\Psi_s(a)$ is a $N$-dimensional vector where the $b$-th wavelet coefficient of $\Psi_s(a)$ represents the information that $v_a$ received from $v_b$. 
Nodes playing similar roles have similar wavelet coefficient. 

\textbf{(3) Characterize by the time-evolution operator. } 
Since the time-evolution operator $U(t)$ reflects the graph state evolution, we treat the wavelets $\Psi$ as probability distributions over graph and use $\phi(\Psi,t)=\mathbb{E}[U(t)]=\mathbb{E}[e^{-i\Psi t}]$ as the characteristic function to uncover nodes' roles in information diffusion. 
The empirical characteristic function of $v_a$ is: 
\begin{equation}
\label{eq:characteristic}
    \phi_s(v_a, t)=\frac{1}{N}\sum^{N}_{n=1}e^{-i\Psi_s(a)t}. 
\end{equation}
$\phi_s(v_a, t)$ can capture all the moments (including higher-order moments) of the given distribution $\Psi_s(a)$. 
We sample at $T$ different time points on the time-evolution operator and then concatenate the values: 
\begin{equation}
\label{eq:h_s_a}
    \mathbf{h}_s(v_a)=\left[{\rm Re}\left(\phi_s\left(v_a, t\right)\right), {\rm Im}\left(\phi_s\left(v_a, t\right)\right)\right]_{t=t_1,t_2,\cdots,T}. 
\end{equation}
In general, the nodes play different roles across different scales. 
Hence we utilize a multi-scale wavelet diffusion strategy to capture both the local and global structural roles of nodes. 
We integrate information across different radii of neighborhoods by jointly considering a set of different values of $s$. 
We can obtain multi-scale structure role encodings $\mathbf{h}_a\in\mathbb{R}^{2TM}$ by concatenating the structural encodings at several different scales $S:\{s_1, s_2, \cdots, s_M\}$: 
\begin{equation}
\label{eq:h_a}
    \mathbf{h}_a=\left (\mathbf{h}_{s_1}\left(v_a\right)|\mathbf{h}_{s_2}\left(v_a\right)|\cdots| \mathbf{h}_{s_M}\left(v_a\right)\right ).  
\end{equation}




\subsubsection{Iterative Structure Learning}
After obtaining the structural role encodings, we use a metric function that accounts for both feature information and the role-based similarities to measure the possibility of edge existence. 
\modelname~is agnostic to various metric functions and we choose the multi-head cosine similarity function here: 
\begin{equation}
\label{eq:metric}
\begin{aligned}
    a^{R(e-1)}_{ij}=\frac{1}{m}\sum^{m}_{h=1}\cos\left(\right.
    &\mathbf{W}_h \cdot \left(\mathbf{z}^{(e-1)}_i|\mathbf{h}^{(e-1)}_i\right), \\
    &\left.\mathbf{W}_h\cdot \left(\mathbf{z}^{(e-1)}_j|\mathbf{h}^{(e-1)}_j\right)\right),     
\end{aligned}
\end{equation}
where $m$ is the number of heads, $\mathbf{W}_h$ is the weight matrix of the $h$-th head, $\mathbf{z}_i^{(e-1)}$ and $\mathbf{h}_i^{R(e-1)}$ denote the representation vector and the structural role encoding vector of node $v_i$ in the $(e-1)$-th epoch, and $|$ denotes the concatenation operation. 
With the above structure learning strategy, we can obtain a role-aware adjacency matrix in the $e$-th epoch: 
\begin{equation}
\label{eq:positionA}
    \mathbf{A}^{(e)}_{R}=\left\{a^{R(e-1)}_{ij}\right\}, i,j\in \{1,2,\cdots,N\}. 
\end{equation}

\subsection{Learning Process of \modelname}
\subsubsection{Dynamic Structure Fusion}
The input graph structure determines the learning performance to a certain extent. 
To avoid the non-convergence or unstable training brought by the poor quality of learned structure at the beginning of training, we hence incorporate the original graph structure $\mathbf{A}$ as supplementary to formulate an optimized graph structure $\Tilde{\mathbf{A}}$: 
\begin{equation}
\label{eq:A*}
    \Tilde{\mathbf{A}}^{(e)}=\gamma\mathbf{D}^{-\frac{1}{2}}\mathbf{A}\mathbf{D}^{-\frac{1}{2}}+\left(1-\gamma\right)\cdot \operatorname{RowNorm}\left(\mathbf{A}^{(e)}_R\right),  
\end{equation}
where $\operatorname{RowNorm}(\cdot)$ denotes the row-wise normalization function, $\gamma$ is a constant that control the contribution of original structure. 
Here we use a dynamic decay mechanism for $\gamma$ to enable the role-aware structure $\mathbf{A}_R^{(e)}$ to play a more and more important role during training. 
Then the refined structure is inputted into a GNN encoder for node representation vectors $\mathbf{Z}\in\mathbb{R}^{N\times d}$ and classification: 
\begin{equation}
    \mathbf{Z}^{(e)}=\operatorname{GNN-Encoder}(\Tilde{\mathbf{A}}^{(e)}, \mathbf{X}). 
\end{equation}

\subsubsection{Objective of \modelname}
The overall loss $\mathcal{L}$ of \modelname~is composed of two terms, the classification loss $\mathcal{L}_{cls}$ and the structure PRI loss $\mathcal{L}_{\rm PRI}$ in Eq.~\eqref{eq:PRI_loss}, given by:
\begin{equation}
\label{eq:loss}
\begin{aligned}
        \mathcal{L}
        &=\mathcal{L}_{cls}+\alpha\mathcal{L}_{\rm PRI}\\
        &=H_{\rm CE}\left(\Tilde{G}, Y\right)+\alpha\left(H_{\rm vN}\left(\Tilde{G}\right)+\beta D_{\rm QJS}\left(\Tilde{G}||G\right)\right),  
\end{aligned}
\end{equation}
where $H_{\rm CE}(\cdot)$ is the cross-entropy loss for classification and $\alpha$ is a hyper-parameter to balance the two loss terms. 
The overall process of \modelname~is shown in Algorithm~\ref{alg:algorithm}. 

\begin{algorithm}[!t]
\caption{The overall process of \modelname~for node classification}
\label{alg:algorithm}
\LinesNumbered
\KwIn{Graph $G$ with node labels $Y$; Number of training epochs $Epochs$; Wavelet scale set $S$; Timepoints $T$;  Hyper-parameters $\alpha$, $\beta$ and $\gamma$. }
\KwOut{Refined graph $\Tilde{G}$; Predicted labels $\hat{Y}$. }
Parameter initialization;\\
\For{$e=1,2,\cdots,Epochs$}{
\tcp{Structural Role Encoding}
\For{$s\in S$}{
    $\Psi^{(e-1)}_s \leftarrow$ Eq.~\eqref{eq:wavelet_s}, $\phi_s(v_a, t) \leftarrow$Eq.~\eqref{eq:characteristic};\\
    $\mathbf{h}^{(e-1)}_s(v_a) \leftarrow$ Eq.~\eqref{eq:h_s_a};\\
}
$\mathbf{h}^{(e-1)}_a \leftarrow$ Eq.~\eqref{eq:h_a};\\
\tcp{Graph Structure Learning}
$a^{R(e)}_{ij} \leftarrow$ Eq.~\eqref{eq:metric}, $\mathbf{A}^{(e)}_R=\left\{a^{R(e)}_{ij}\right\}$;\\
$\Tilde{\mathbf{A}}^{(e)} \leftarrow $Eq.~\eqref{eq:A*}, $\Tilde{G}^{(e)}  \leftarrow (\mathbf{X}, \Tilde{\mathbf{A}}^{(e)})$;\\
\tcp{Learn Node Representations}
$\mathbf{Z}^{(e)}=\operatorname{GNN-Encoder}(\Tilde{\mathbf{A}}^{(e)}, \mathbf{X})$;\\
\tcp{Optimize}
$\mathcal{L}^{(e)}_{\rm PRI}=H_{\rm vN}(\Tilde{G}^{(e)})+\beta D_{\rm QJS}(\Tilde{G}^{(e)}||G)$;\\
$\mathcal{L}^{(e)}=\mathcal{L}^{(e)}_{cls}(\Tilde{G}^{(e)}, Y)+\alpha\mathcal{L}^{(e)}_{\rm PRI}$; \\
Update model parameters to minimize $\mathcal{L}^{(e)}$.
}
\end{algorithm}

\subsubsection{Approximation}
Recall that the computation of von Neumann entropy and spectral wavelets requires the full eigenvalue decomposition of the Laplacian matrix, which takes $O(N^3)$ time. 
The von Neumann entropy can be approximated with linear complexity $O(|V|+|E|)$~\cite{chen2019fast}. 
In the experiments, we still use the basic von Neumann entropy. 
As for the graph wavelets, we use the Chebyshev polynomial approximation~\cite{shuman2011chebyshev} to compute $\Psi$, reducing the computational complexity to $O(K|E|)$, where $K$ is the order of Chebyshev polynomials. 


\subsection{Properties of $\Tilde{G}$ Learned by \modelname}
\label{sec:property}
\modelname~provides a unified way to control the quality of learned graph structure in terms of \textit{sparsity, centrality, and nuisance invariance property}. 
\subsubsection{Sparsity and Centrality}
The von Neumann entropy has close connections with the structure sparsity and centrality. 
As indicated in~\cite{passerini2008neumann}, given a graph $G$, let $G'=G+\{x,y\}$ with $V(G')=V(G)$ and $E(G')=E(G)\cup\{x,y\}$, then $H_{\rm vN}(\rho_{G'})\ge \frac{d_{G'}-2}{d_{G'}}H_{\rm vN}(\rho_{G})$. 
The von Neumann entropy tends to grow with the increasing number of edges. 
The graph centrality (i.e., the extent to which a graph is organized around some central nodes) can be measured as a quantum relative entropy between the relative degree distribution and the uniform distribution~\cite{simmons2018neumann}, which is given by $D(\rho_G||I_N):=\textbf{tr}(\rho_G (\log \rho_G - \log I/N))=\log N-H_{\rm vN}(G)$. 
The von Neumann entropy tends to grow with the increasing regularity of the graph. 
The above conclusions suggest that minimizing $H_{\rm vN}(\Tilde{G})$ leads to a sparse and centralized structure. 



\subsubsection{Nuisance Invariance}
$\Tilde{G}$ only preserves the most relevant yet least redundant information in the observed graph $G$ and is invariant to nuisances in data. 
Suppose $G_n\in G$ the task-irrelevant nuisance in $G$, the relevance of $\Tilde{G}$ and $G_n$ can be formulated as the divergence between their conditional distributions predictions of the desired labels $Y$~\cite{yu2020measuring}: $\mathbb{E}\left[D_{\rm KL}\left(p\left(Y|\Tilde{G}\right)\right)||p\left(Y|G_n\right)\right]$. 
Since $G_n$ is irrelevant with $Y$, we have $p(Y|G_n)=p(Y)$. 
Minimizing the cross-entropy loss (i.e., the mutual information $I(Y;\Tilde{G})$ between $\Tilde{G}$ and $Y$) is equivalent to minimizing the relevance between $\Tilde{G}$ and $G_n$: 
\begin{equation}
\begin{aligned}
    &\mathbb{E}\left[D_{\rm KL}\left(p\left(Y|\Tilde{G}\right)\right)||p\left(Y|G_n\right)\right]\\
    &=\mathbb{E}\left[D_{\rm KL}\left(p\left(Y|\Tilde{G}\right)\right)||p\left(Y\right)\right]\\
    &=\iint\left(p\left(Y|\Tilde{G}\right)\log\frac{p\left(Y|\Tilde{G}\right)}{p\left(Y\right)}\right)p\left(\Tilde{G}\right)\\
    &=\iint p\left(Y|\Tilde{G}\right)\log\frac{p\left(Y,\Tilde{G}\right)}{p\left(Y\right)p\left(\Tilde{G}\right)}\\
    &=I(Y;\Tilde{G}).
\end{aligned}
\end{equation}


\section{Experiments}
\label{sec:exp}
We evaluate \modelname~on node classification and graph denoising tasks
to verify its capability of improving the effectiveness and robustness of graph representation learning. 
Then we provide the analyses of the PRI loss, the structural role encodings, and the learned structure.

\begin{table*}[!htp]
\caption{Accuracy ± standard deviation (\%) of node classification.  (\textbf{Bold}: best result; {\ul Underlined}: runner up. ) }
\label{tab:node_cls}
\centering
\begin{tabular}{cccccccc}
\hline
\textbf{Method} &
  \textbf{\begin{tabular}[c]{@{}c@{}}Squirrel\\ $h$=0.22\end{tabular}} &
  \textbf{\begin{tabular}[c]{@{}c@{}}Actor\\ $h$=0.24\end{tabular}} &
  \textbf{\begin{tabular}[c]{@{}c@{}}Chameleon\\ $h$=0.25\end{tabular}} &
  \textbf{\begin{tabular}[c]{@{}c@{}}CiteSeer\\ $h$=0.72\end{tabular}} &
  \textbf{\begin{tabular}[c]{@{}c@{}}PubMed\\ $h$=0.79\end{tabular}} &
  \textbf{\begin{tabular}[c]{@{}c@{}}Cora\\ $h$=0.83\end{tabular}} &
  \textbf{\begin{tabular}[c]{@{}c@{}}Photo\\ $h$=0.83\end{tabular}} \\ \hline
GCN           & 22.22±1.24          & 22.85±1.64          & 32.16±2.76          & 66.31±1.12       & 74.11±3.65         & 79.10±0.77          & 85.61±2.20          \\
GAT           & 22.64±1.25          & 23.53±1.33          & 32.05±2.56          & 63.85±2.45       & 73.02±2.52         & 77.86±1.48          & 87.97±2.73          \\
GraphSAGE     & 28.79±1.74          & 24.22±1.44          & 37.05±2.35          & 64.80±1.83       & 72.61±2.95         & 75.23±1.31          & 86.23±2.53          \\ \hline
DropEdge      & 22.35±1.12          & 23.84±1.20          & 32.62±2.75          & 66.68±1.38       & 75.97±0.82         & 79.30±0.84          & 86.05±1.78          \\
NeuralSparse  & 29.02±1.10          & 24.50±1.42          & 47.30±2.22          & 67.82±1.18          & 74.87±2.77      & 81.47±1.43          & {\ul 89.40±1.85}    \\
Graph-PRI     & 28.44±2.10          & 23.81±2.31          & 42.39±1.99          & 69.24±1.25       & {\ul 76.25±1.44}         & 79.07±1.12          & 88.30±2.11          \\ \hline
IDGL          & {\ul 29.13±2.94}    & {\ul 27.44±5.80}    & {\ul 49.80±4.80}    & 67.94±0.28       & 75.32±1.45         & {\ul 83.23±0.62}    & 88.89±2.55          \\
Pro-GNN       & 27.18±1.28          & 24.82±2.81          & 48.54±4.87          & 66.68±2.02       & 75.44±3.54         & 82.14±0.58          & 87.28±1.85          \\
SDRF          & \textgreater 1 day  & \textgreater 1 day  & 41.05±1.17          & \textbf{69.97±0.28} & \textgreater 1 day & 81.94±0.59          & \textgreater 1 day  \\
SLAPS         & 25.29±1.06          & 23.10±3.39           & 40.24±1.80          & 68.58±1.46       & 75.64±0.77         & 79.27±1.54          & 88.48±2.47          \\
\textbf{\modelname} & \textbf{33.87±2.08} & \textbf{28.55±2.04} & \textbf{51.83±2.44} & {\ul 69.34±2.64}      & \textbf{76.77±3.20}      & \textbf{83.67±2.09} & \textbf{92.40±1.15} \\ \hline
\end{tabular}
\end{table*}
\subsection{Experimental Settings}
\subsubsection{Datasets}
We select datasets with different homophily ratios $h$~\cite{pei2020geom} to analyze methods' generalization on graphs with different properties. 
The evaluation datasets are Squirrel, Chameleon~\cite{rozemberczki2021multi}, Actor~\cite{pei2020geom}, CiteSeer, PubMed, Cora~\cite{sen2008cora} and Photo~\cite{shchur2018pitfalls}. 

\subsubsection{Baselines}
We consider three types of baselines: 
(1) \textit{Graph neural networks}: GCN~\cite{kipf2016semi}, GAT~\cite{velivckovic2017graph} and GraphSAGE~\cite{hamilton2017inductive}; 
(2) \textit{Graph sparsification methods}: DropEdge~\cite{rong2019dropedge}, NeuralSparse~\cite{zheng2020robust}, 
and Graph-PRI~\cite{yu2022principle}; 
(3) \textit{Graph structure learning methods}: IDGL~\cite{chen2020iterative}, Pro-GNN~\cite{jin2020graph}, SDRF~\cite{topping2021understanding}, and SLAPS~\cite{fatemi2021slaps}. 

\subsubsection{Parameter Settings}

We re-implement the NeuralSparse~\cite{zheng2020robust} and SDRF~\cite{topping2021understanding}. 
The parameters of baseline methods are set to the suggested value in their papers or carefully tuned for fairness. 
For the GNN encoders, we use a 2-layer GCN for node classification.  
We set the representation dimension $d$=32, the Chebyshev polynomial order $K$=10, the number of time points $T$=4, the number of scales $M$=2, and the number of heads $m$=4. 
The other hyper-parameters ($\alpha$, $\beta$, and $\gamma$) are tuned for each dataset. 
\begin{figure}
    \centering
    \includegraphics[width=\linewidth]{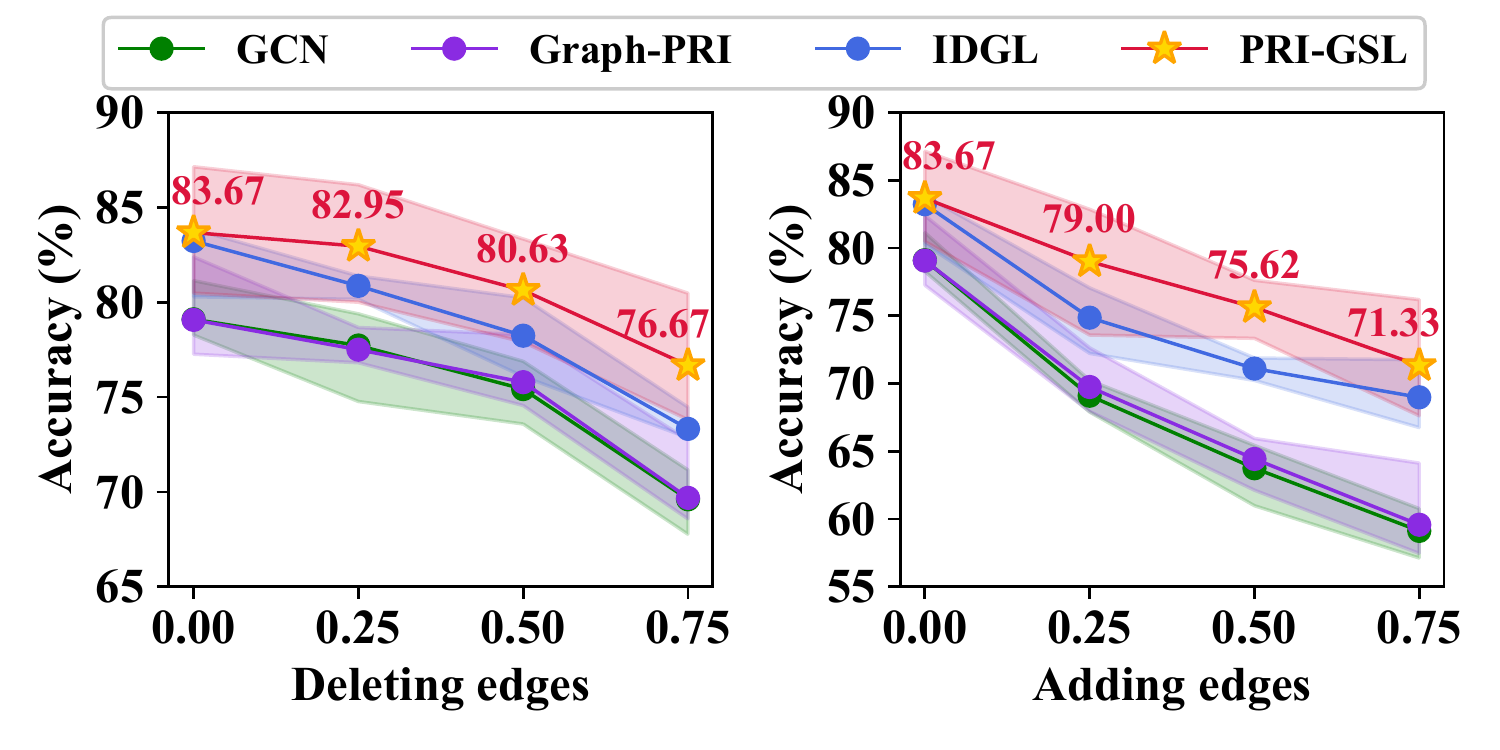}
    \caption{\modelname~on noisy graphs. }
    \label{fig:denoise}
\end{figure}
\subsection{Evaluation Results and Analysis}
\subsubsection{Node Classification}
We set the number of nodes in each class to be 20/30 for training/validation and take the remaining nodes for the test. 
The accuracy and standard deviation on 10 randomly split are shown in Table~\ref{tab:node_cls}. 
The best results are shown in bold and the runner-ups are underlined. 
Our \modelname~achieves the best performance on both homophilic and heterophilic datasets, showing the effectiveness of utilizing the self-organization property when mining the latent inherent structure. 
Generally, the graph structure learning methods show better performance than GNNs and graph sparsification methods. 
Pro-GNN and SLAPS perform well on the homophilic graphs but show unsatisfactory performance on the heterophilic graphs (Actor, Chameleon, and Squirrel), demonstrating the limitation of using heuristic assumptions for structural constraints. 
Although Graph-PRI also uses PRI for graph sparsification, it achieves fewer improvements since it does not take the node feature and the downstream task into consideration.

\begin{figure}
    \centering
    \includegraphics[width=\linewidth]{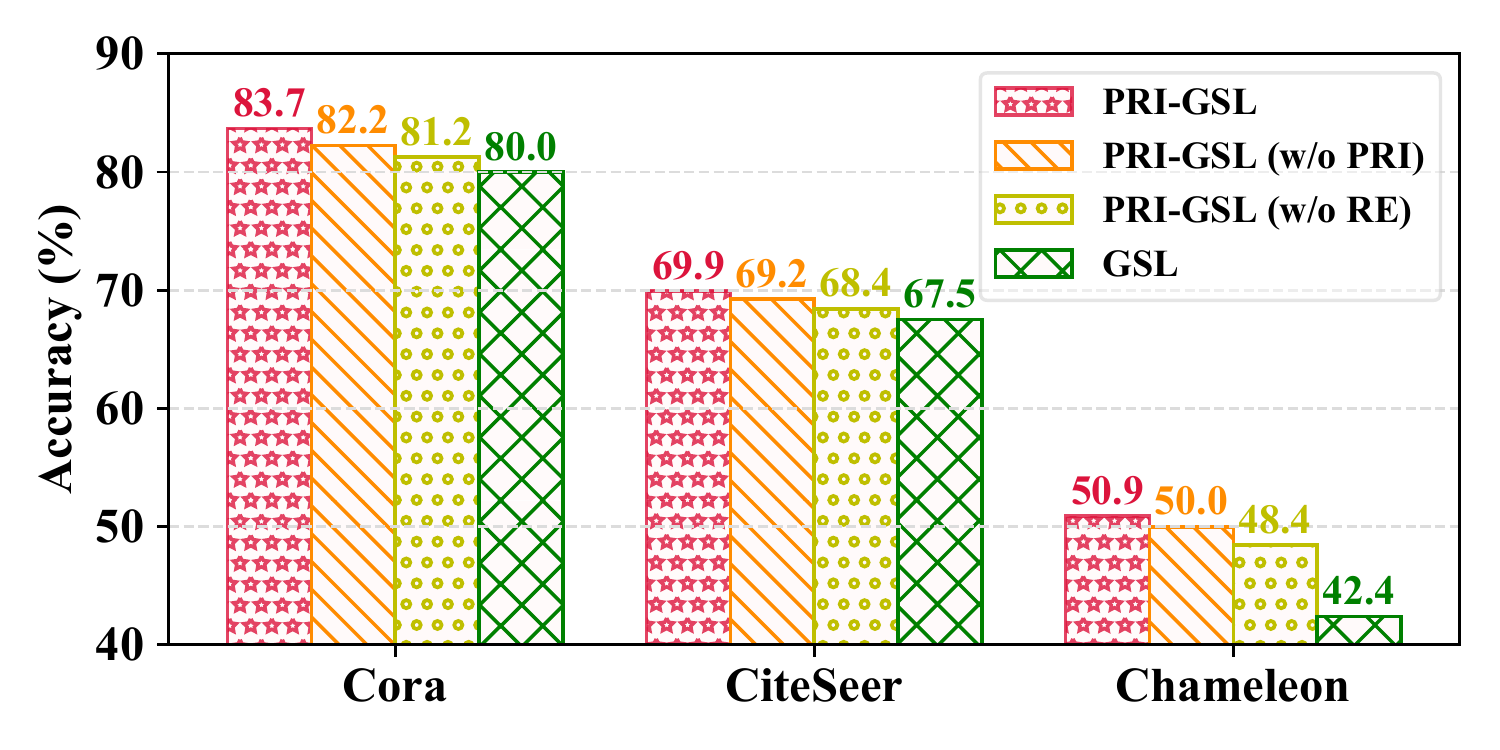}
    \caption{Ablation study results of \modelname. }
    \label{fig:abla}
\end{figure}
\begin{figure*}[!h]
\centering
\subfigure[{Original Graph. }]{
\begin{minipage}[t]{ 0.185\linewidth}
\centering
\includegraphics[width=\linewidth]{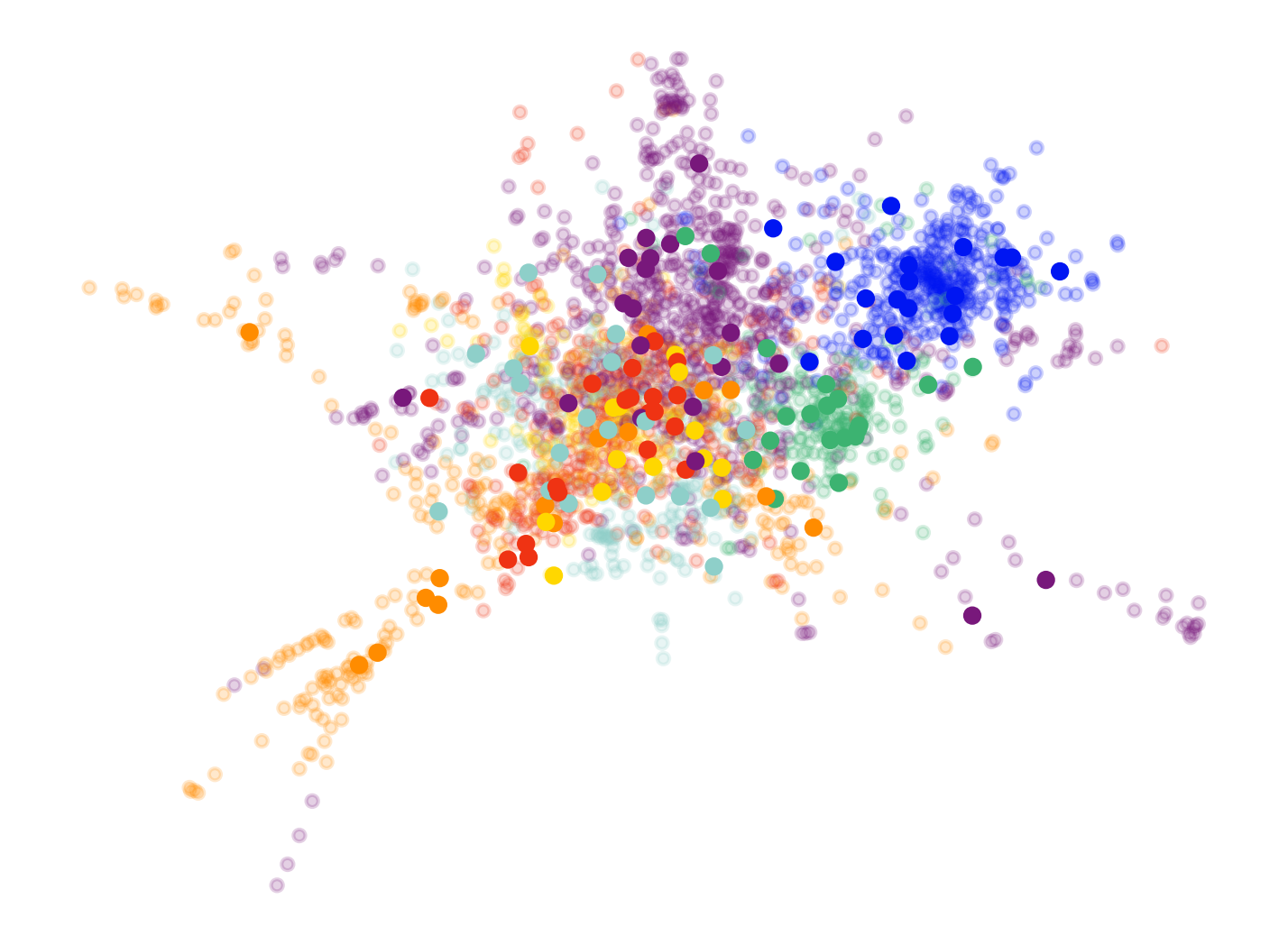}
\label{fig:cora}
\end{minipage}
}
\subfigure[Graph-PRI. ]{
\begin{minipage}[t]{ 0.185\linewidth}
\centering
\includegraphics[width=\linewidth]{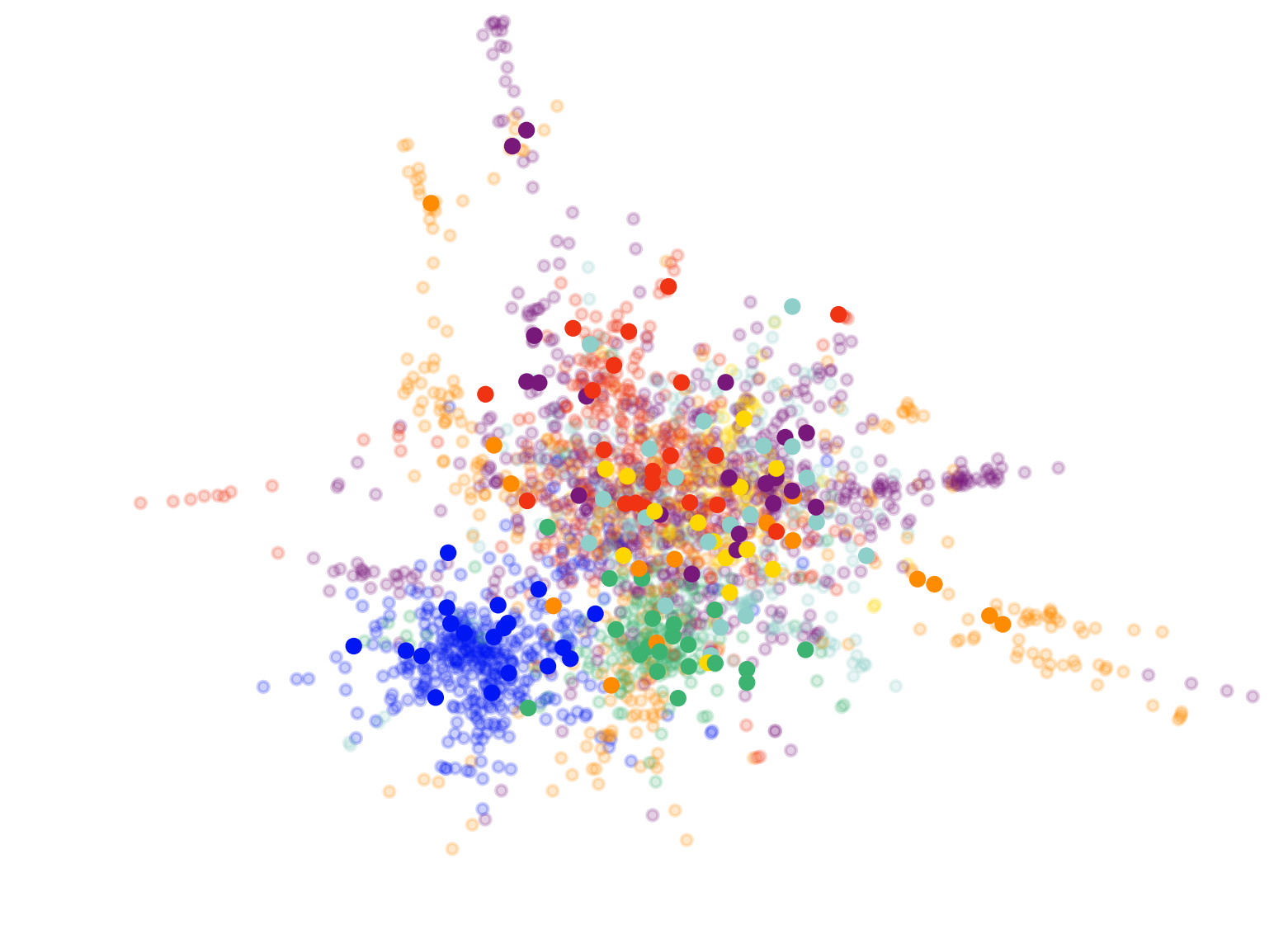}
\label{fig:renode}
\end{minipage}
}
\subfigure[Pro-GNN. ]{
\begin{minipage}[t]{ 0.185\linewidth}
\centering
\includegraphics[width=\linewidth]{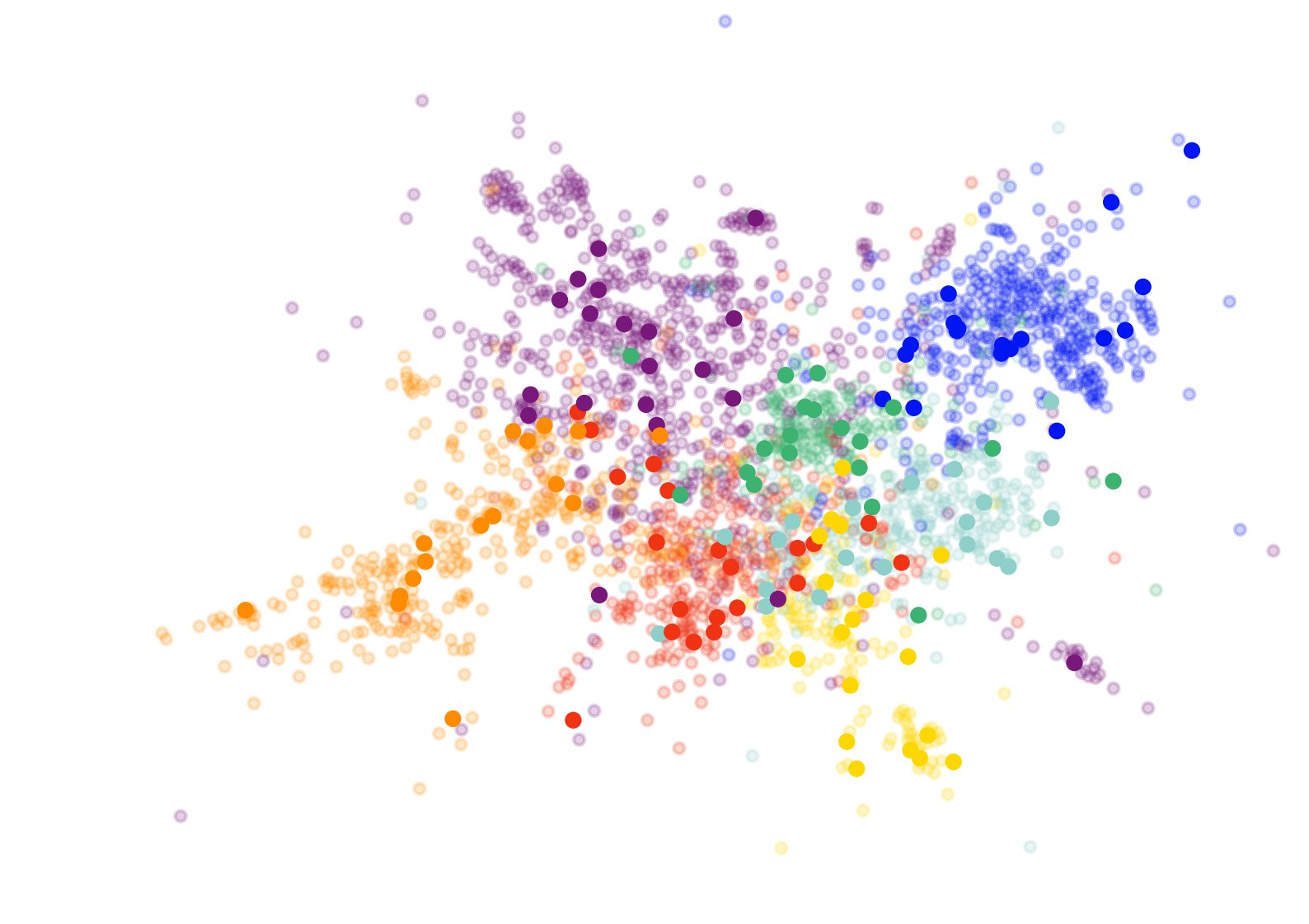}
\label{fig:SDRF}
\end{minipage}
}
\subfigure[IDGL. ]{
\begin{minipage}[t]{ 0.185\linewidth}
\centering
\includegraphics[width=\linewidth]{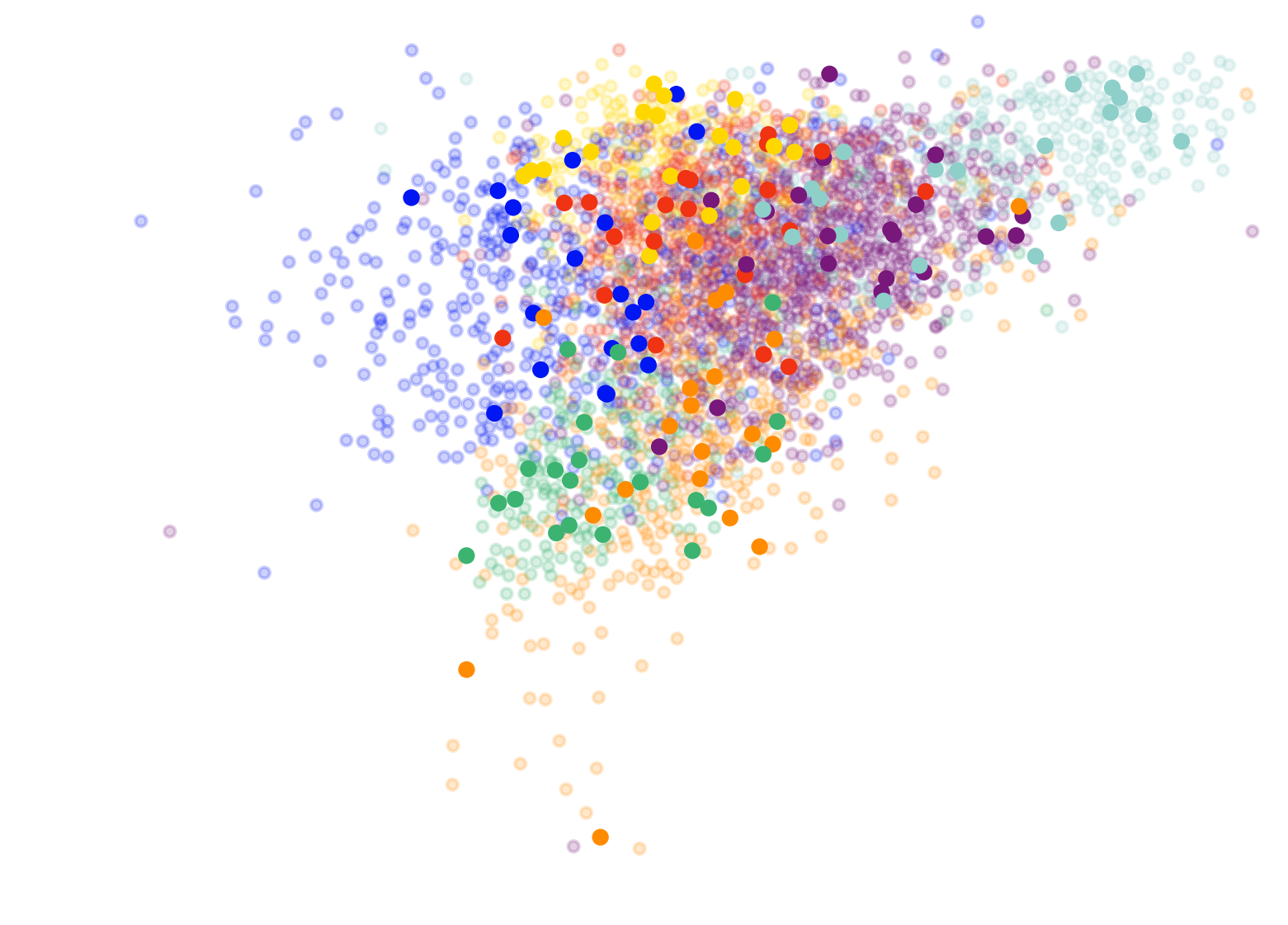}
\label{fig:IDGL}
\end{minipage}
}
\subfigure[\modelname. ]{
\begin{minipage}[t]{ 0.185\linewidth}
\centering
\includegraphics[width=\linewidth]{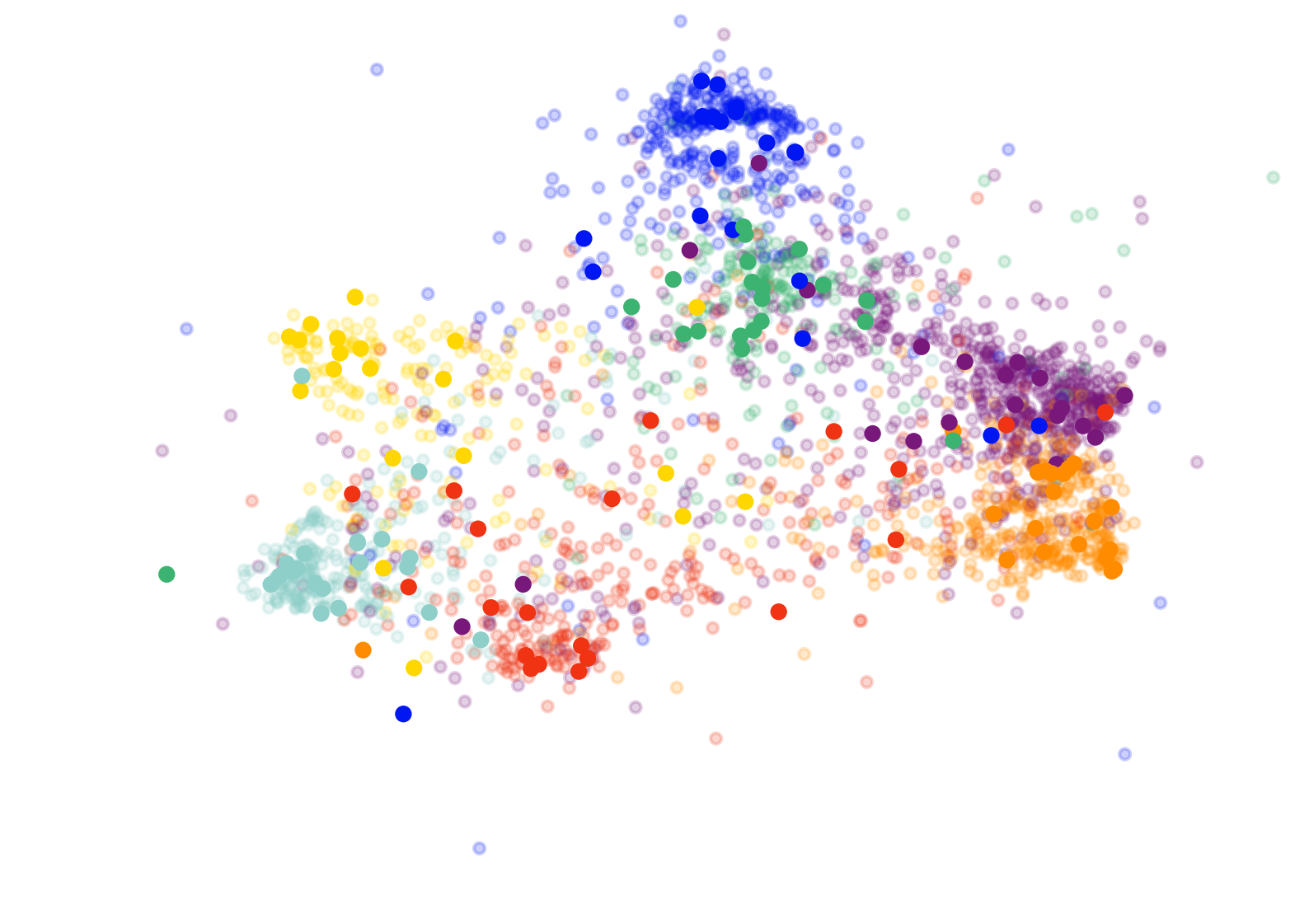}
\label{fig:ours}
\end{minipage}
}
\centering
\caption{Visualization of the original graph of Cora and learned graphs by Graph-PRI, Pro-GNN, IDGL, and \modelname. }
\label{fig:visual}
\end{figure*}

\subsubsection{Graph Denoising}
To evaluate \modelname's ability to move noisy information, we generate synthetics noisy datasets by adding/deleting edges on Cora following~\cite{chen2020iterative,sun2022graph}. 
Specifically, we randomly add/delete 25\%, 50\%, 75\% edges for 5 times and show the mean accuracy (solid line) and standard deviation (shaded region) in Fig.~\ref{fig:denoise}. 
The performance of GCN and Graph-PRI decreases dramatically with the increasing noise level. 
The structure learning methods, IDGL and \modelname, show more robustness compared to vanilla GCN and Graph-PRI. 
Adding edge hurts more than deleting edges, indicating the importance of removing redundant information in the structure. 
\modelname~consistently shows better performance under different levels of external noise than the other baselines. 
Even though Graph-PRI shares the same spirit with \modelname, it fails to distinguish whether the noise is task-relevant and shows the same poor robustness as GCN. 
The performance of \modelname~on noisy graphs also demonstrates the nuisance invariance property.  


\subsubsection{Ablation Study}
\label{sec:abla}
To illustrate the advantages of the guidance of PRI and structural role information, we compare \modelname~with three variants: 
(1) \modelname~(w/o PRI) that removes the PRI loss, (2) \modelname~(w/o RE) that removes the role encodings, and 
(3) GSL that removes both the PRI loss and the role encoding. 
The results of variants on 5 random split datasets are shown in Fig.~\ref{fig:abla}. 
As we can observe, both the PRI loss and the structural role encoding benefit the classification, where the structural role encoding brings more improvement. 
This suggests that it's important to capture the node's contribution to the graph information flow when identifying the graph organization. 

\begin{figure}[t]
    \centering
    \includegraphics[width=\linewidth]{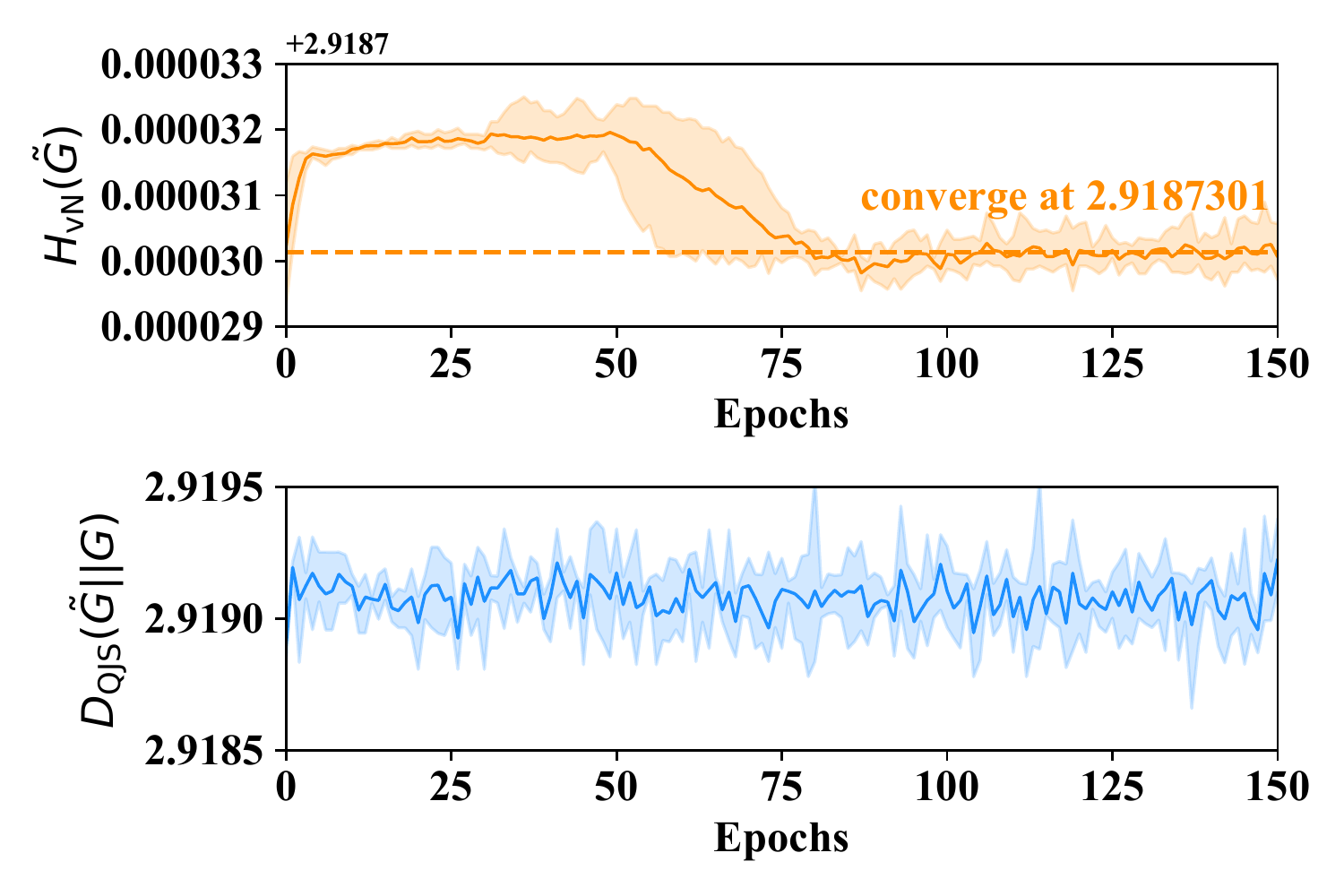}
    \caption{The variations of $H_{\rm vN}(\Tilde{G})$ and $D_{\rm QJS}(\Tilde{G}||G)$. }
    \label{fig:entropy_diver}
\end{figure}

\subsubsection{Empirical Behavior of $H_{\rm vN}(\Tilde{G})$ and $D_{\rm QJS}(\Tilde{G}||G)$}
We analyze the learning dynamics of \modelname~by measuring the variations of $H_{\rm vN}(\Tilde{G})$ and $D_{\rm QJS}(\Tilde{G}||G)$ on Cora with $\alpha$=0.1 and $\beta$=1 in Fig.~\ref{fig:entropy_diver}. 
The shadowed area is enclosed by the min and max value of four training runs. 
The solid line in the middle is the mean value of each epoch. 
$H_{\rm vN}(\Tilde{G})$ first increases for about 50 epochs with the structure exploration and then decreases to converge after about the 80-th epoch, indicating that the learned structure is with high certainty. 
$D_{\rm QJS}(\Tilde{G}||G)$ bumps during the training process. 
This may be because the model continues to seek a balance of structure redundancy and distortion during the training process. 


\subsubsection{Hyper-parameter Analysis}
\begin{figure}[t]
    \centering
    \includegraphics[width=\linewidth]{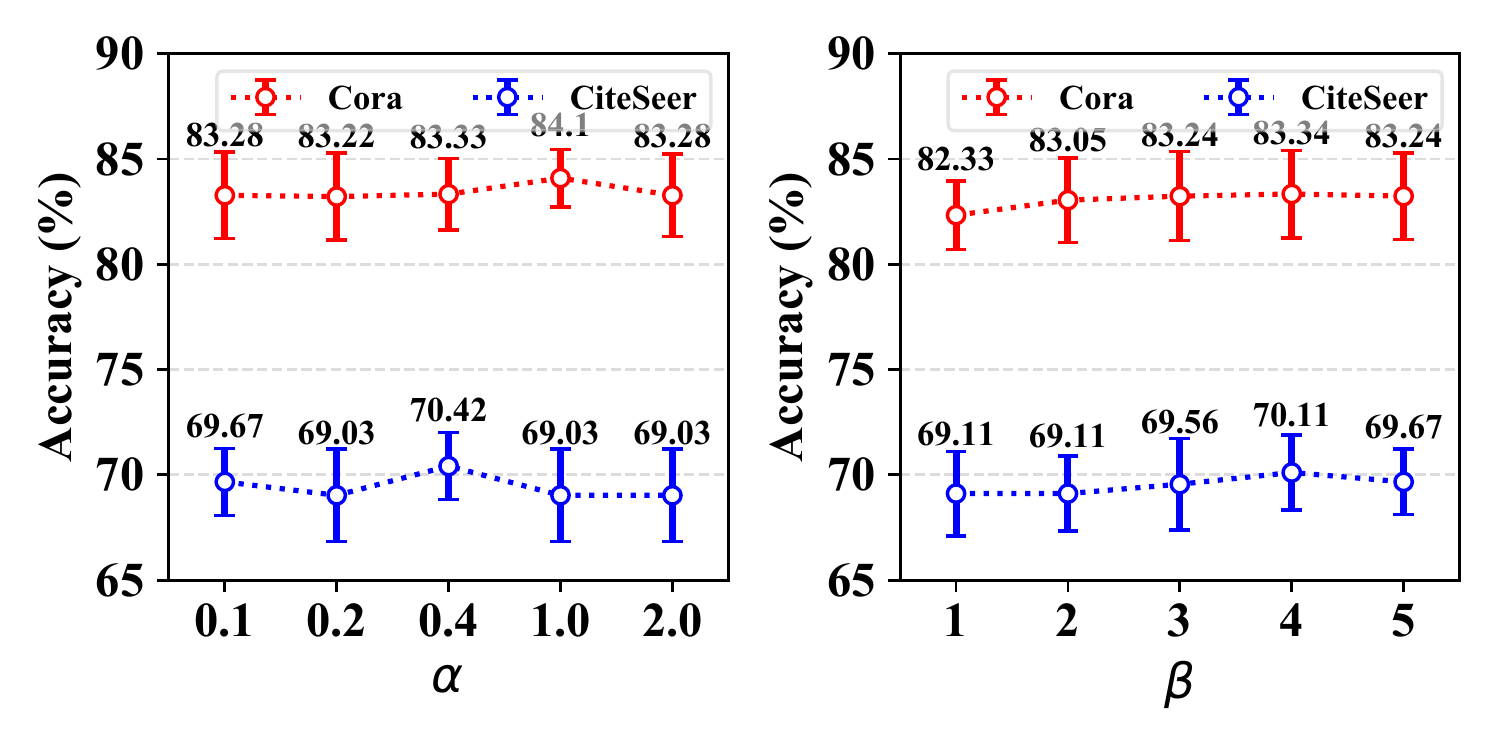}
    \caption{Parameter sensitivity of $\alpha$ and $\beta$. }
    \label{fig:sensitivity}
\end{figure}

We analyze the impact of hyper-parameters including $\alpha$ controlling the importance of the PRI loss in Eq.~\eqref{eq:loss} and $\beta$ trading off redundancy and distortion in Eq.~\eqref{eq:PRI_loss}. 
The results are shown in Fig.~\ref{fig:sensitivity}. 
\modelname~achieves the best performance with $\alpha$=1 on Cora and $\alpha$=0.4 on CiteSeer, which indicates that \modelname~benefits from the PRI loss. 
As for the $\beta$ in the PRI loss, when the distortion term has a weight more than 2 compared to the redundancy term, \modelname~could reach satisfactory performance on both datasets. 
This suggests that the distortion term dominates the PRI loss in \modelname. 
That is to say, the learned structure should preserve enough information from the original graph to perform well on the downstream task. 

\subsection{Visualization}
In Figure~\ref{fig:visual}, we visualize the original graph structure of Cora and the graphs learned by PRI-Graph, Pro-GNN, IDGL, and \modelname~using \textit{networkx}. 
The nodes' colors indicate their classes, the labeled nodes are solid and the unlabeled nodes are hollow. 
The edges are not shown for clarity and the layout of nodes represents their connectivities. 
Graph-PRI has little effect on the overall property of graph structure. 
Even though Pro-GNN and IDGL can make nodes within different classes more separate, there are still some overlapping and entangled areas. 
Benefiting from the structural role encoding, \modelname~can obtain the structure with separate clusters with similar shapes and clearer class boundaries, showing how the nodes within a class are organized. 

\section{Conclusion}
In this work, we propose an information-theoretic framework for graph structure learning named \modelname. 
We formulate the Principle of Relevant Information for graph data to quantify the structure redundancy and distortion, which acts as a potentially unifying guidance of structure learning. 
We propose a role-aware structure learner based on the quantum continuous walk evolution to unravel the self-organization of the graph. 
The learned structure enjoys the property of sparse, centralized, and nuisance invariance. 
Extensive experiments demonstrate the
superior effectiveness and robustness of \modelname. 
\clearpage
\section{Acknowledgments}
The corresponding author is Jianxin Li. 
The authors are supported by the NSFC through grant No.U20B2053, 
and in part by NSF under grants III-1763325, III-1909323, III-2106758, and SaTC-1930941.

\bibliography{ref}

\end{document}